\RequirePackage{amsmath}
\documentclass[a4paper]{article}
\usepackage[utf8]{inputenc}
\usepackage[T1]{fontenc}
\usepackage{adjustbox,lipsum}

\usepackage[a4paper,top=3cm,bottom=2cm,left=3cm,right=3cm,marginparwidth=1.75cm]{geometry}

\usepackage{amsmath}
\usepackage{graphicx}
\usepackage{algorithm}
\usepackage{algpseudocode}
\usepackage{multirow}
\usepackage{todonotes}

\usepackage[colorlinks=true, allcolors=blue]{hyperref}

\begin{document}

\title{Real-time Bidding campaigns optimization using attribute selection}
\author{Luis Miralles-Pechuan, M. Atif Qureshi, and Brian Mac Namee}
\date{%
Centre for Applied Data Analytics Research (CeADAR), University College Dublin, Dublin, Ireland.
    \today
}

\maketitle
\begin{abstract}
Real-Time Bidding is nowadays one of the most promising systems in the online advertising ecosystem. In the presented study, the performance of RTB campaigns is improved by optimising the parameters of the users' profiles and the publishers' websites. Most studies about optimising RTB campaigns are focused on the bidding strategy. In contrast, the objective of our research consists of optimising RTB campaigns by finding out configurations that maximise both the number of impressions and their average profitability. An online campaign configuration generally consists of a set of parameters along with their values such as \{Browser = "Chrome", Country = "Germany", Age = "20-40" and Gender = "Woman"\}. Throughout the investigation, a series of experiments have been conducted in order to verify that it is possible to increase the performance of RTB campaigns by means of configuration optimisation.
The experiments demonstrate that, when the number of required visits by advertisers is low, it is easy to find configurations with high average profitability, but as the required number of visits increases, the average profitability tends to go down. Additionally, configuration optimisation has been combined with other interesting strategies to increase, even more, the campaigns' profitability. Along with parameter configuration the study considers the following complementary strategies to increase profitability: i) selecting multiple configurations with a small number of visits instead of a unique configuration with a large number, ii) discarding visits according to the thresholds of cost and profitability, iii) analysing a reduced space of the dataset and extrapolating the solution, and iv) increasing the search space by including solutions below the required number of visits. The developed campaign optimisation methodology could be offered by RTB platforms to advertisers to make their campaigns more profitable.
\end{abstract}
\section{Introduction}

The emergence of big data technologies and the facility to store a large volume of data has allowed companies to extract valuable information for their businesses \cite{provost2013data,grochowski1996future}. In particular, in the online advertising business, companies collect information about users traffic and websites performance to maximise the benefits of advertisers, publishers and advertising networks (AN). Additionally, given the ease of collecting data, online advertising has become an active area of research where machine learning techniques are frequently used to predict events related to the campaign’s performance such as the probability of generating a click in a given advert \cite{wang2017display}. 

The first online publicity banner appeared in October 1994 and it belonged to a company called Hotwired. Ever since the online publicity sector has been growing as the number of active users on the Internet increased. The amount of people with access to the Internet makes online marketing, an attractive choice compared to traditional marketing channels. Some of the companies that started in this niche market around two decades ago such as Google and Facebook, are nowadays among the most important companies in the world \cite{goldfarb2014different}.

In the beginning, the ecosystem of online advertising was very simple; advertisers negotiated a price for displaying an advert directly with a publisher. For each campaign, it was required negotiation of the price which made the ecosystem unscalable as the demand of online adverts grew. Furthermore, the minimum investment required was a barrier to small businesses. In order to overcome the drawbacks of scalability, a new ecosystem called Advertising Networks (AN) was proposed \cite{zeff1999advertising}.

The AN acts as an intermediary between advertisers and publishers. The main tasks of ANs are: buying and selling impressions, displaying adverts on the publishers' websites, and preventing fraud on the publicity ecosystem. Generally, ANs empowers advertisers with online tools to facilitate managing the advertising process and to inform them about the yield of their campaigns almost in real-time. In the same way, those tools inform publishers in detail about the performance of their websites \cite{trattner2013social}.

ANs have been a popular choice for some time but, a few years ago, a new sophisticated auction-based model called Real-time bidding (RTB) emerged. RTB offers important advantages over the AN model; it connects directly advertisers with publishers, eliminating intermediaries and their respective commissions, and it increases the cost-benefit of both sides.

RTB is a real-time auction platform, where buying and selling online impressions takes place instantaneously on a per-impression basis via programmable criteria. RTB has a hierarchical bidding system where impressions are offered through several subsystems and the highest price is selected. When a bid is won by an advertiser, the advert is displayed on the publisher's website \cite{wang2017display}. These bidding and winning processes are executed within the time it takes a user to load a page (100 milliseconds). Additionally, RTB is able to target audiences based on interests and profiles by examining the cookies generated by the users \cite{yuan2014survey}.

RTB offers more transparency to the publisher since they can see the price the advertiser is willing to pay. RTB also gives publishers a higher control over the advert they want to display by allowing them to define a list of banned advertisers. On the other hand, advertisers can buy impressions individually in a fine-grained fashion, which leads to more effective campaigns \cite{yuan2013real}. Moreover, RTB also allows publishers to connect with multiple ANs and advertising agencies (companies that help advertisers to create and manage publicity campaigns), making it a heterogeneous ecosystem. Finally, RTB eliminates the need for commissions or fees to ANs by allowing advertisers to directly buy individual impressions from the publisher \cite{yuan2013real}.

To guaranty the adequate operation of the RTB system, it is required to collect and analyse information of the involved roles on a regular basis. Besides that, the data needs to be processed and converted into actionable information to maximize the performance of online campaigns and to mitigate online fraud \cite{mcmahan2013ad,menon2011response}. \textit{Machine Learning (ML}) techniques are a perfect approach to address those challenges and are widely applied in the domain of online advertising \cite{berry1997data}. 

ML models are trained using features of the users profile (browser, operating system, access time, device kind, or type of access), and of the publishers' websites (position of the banner, category, language, or history of the page) to accurately estimate relevant aspects related to the advertising ecosystem such as Click-through rate (CTR) or conversion rate \cite{mcmahan2013ad}. ML methods generate models that have been successfully applied to predict the CTR of an advert, to estimate the probability of generating a conversion (a purchase, a phone call, filling out a form), and to calculate the right value for a bid \cite{mcmahan2013ad}.

In recent years, there have been many publications related to online campaigns optimisation \cite{zhang2017managing,yuan2014survey}. However, some optimisation techniques are more suitable than others depending on the digital marketing format (banners, text adverts, search engine marketing...) and where advertising campaigns are launched (web searchers, websites, apps...). For example, an effective technique in search engine marketing consists of selecting multiple less frequent and therefore cheaper keywords, but that the sum of them add up a larger number than that of more popular keywords \cite{abhishek2007keyword}. This strategy, however, cannot be applied in RTB because advertisers do not bid based on keywords but on single impressions. In contrast, it is possible to apply the proposal of Thomaidou et al. \cite{thomaidou2012ad}, where an expert human is replaced by a system to save costs, to most digital marketing formats. The proposed system is able to launch, monitor, and configure a search engines campaign automatically.

Regardless of the underlying platform, a key aspect of any successful campaign is to display the advertised product to the right target \cite{sivadas1998internet}, and RTB, as well as ANs, allows advertisers to target specific audiences which are assumed to become receptive towards the marketing of a product. The strategy to aim campaigns towards a small group with common interests is called microtargeting \textbf{(CITATION)}. In addition to that, behavioural targeting, where the user interest is considered to display a relevant advert has been shown to be effective \cite{yuan2013real}. However, making appropriate matchmaking is not a simple task, and even small improvements, when are applied to billions of transactions per day, it turns out into a big amount of money.

Generally, there are two types of campaigns. The first type is called brand-based campaigns and is aimed at maximising the overall long-term revenue \cite{broussard2000advertising}. Brand-based campaigns objective is focused on establishing or increasing brand reputation. The second type is called performance-based campaigns and is aimed at maximising the short-term revenue \cite{chen2011real}. Their objective is selling a product to generate more profits. In our research, we are focused on the last one, performance-based campaigns.

In this paper, we propose a methodology to optimise advertising campaigns in RTB platforms by configuring parameters for both users (to whom the campaign is directed) and web pages (where adverts are displayed). An example of suggested configuration to an advertisers could be: \{Browser = "Chrome", Device OS = "Android",  Traffic Category = "Organic search", Age = "25-34", Gender = "Male", Time = "10:00-11:00", Country = "UK", Position banner = "Top", and Website Category = "Fashion"\}.

The major contribution of the proposed work is developing a method to suggest advertisers with configurations that increase their campaign profitability. Our method is focused on optimising the attribute selection instead of optimising the bidding strategy. To ensure the stated aim, the methodology explores all the possible configurations and ranks them according to the number of visits that meet the requirements and their average profitability. The profitability of an impression is calculated according to the price and the probability that a conversion is generated from the impression. A supervised logistic regression method is applied to estimate the profitability of the advert impressions. To the best of our knowledge, this is the first work in RTB in this direction.

The rest of the paper is organised as follows: Section 2 presents a review of the current state-of-the-art in RTB. In section 3 is described the methodology used to optimise campaigns using parameter selection. In section 4, experiments are conducted to evaluate the different strategies for improving optimization based on parameters. Finally, in section 5, it is presented the conclusions and some directions for future works.

\section{Background on Real-time Bidding and Online campaign optimization}

In this section, it is presented the state-of-the-art of RTB optimisation. First, it is introduced a general overview of online advertising campaigns, subsequently optimisation techniques applied to RTB campaigns are discussed, and finally, some research about RTB campaigns optimisation based on parameters which is the niche targeted in this contribution is described.

\subsection{The general overview of the online advertising campaigns}

In online advertising campaigns, advertisers try to acquire enough number of visits to fulfil their target at the lowest possible price, and that can be achieved using different kinds of platforms such as search engines, blogs, or social networks. There are different strategies to target the right audience such as selecting relevant keywords on a search engine or setting up relevant parameters related to users and website attributes \cite{evans2009online}. Most importantly, a key aspect of a successful campaign lies in the way it is defined (i.e., target, message, adverts and channel).

The series of questions while defining an online advertising campaign are the same as for traditional advertising campaign. Questions such as: Who is the right audience and their demographics? What shall be the design and message that would appeal (e.g., emotional, rational) audience in the right way? Which platform should be used to reach out to the audience effectively? and How to plan an effective campaign within the allocated budget? \cite{sivadas1998internet,yuan2013real}. Consequently, the success of the campaign is measured in terms of the impact i.e., Did campaign met the defined objectives within the budget? 

Online advertising is also applied in the area of predictive models such as personalised systems, user behaviour modelling, or CTR estimation \cite{wang2017display}. Due to monetary benefits, online advertising has become a target of cybercrime and online frauds \cite{daswani2008online}, and consequently, there are many works addressing fraud prevention such as detecting bots simulating real users \cite{laleh2009taxonomy}.

\begin{figure}
\centering
\includegraphics[scale=0.5]{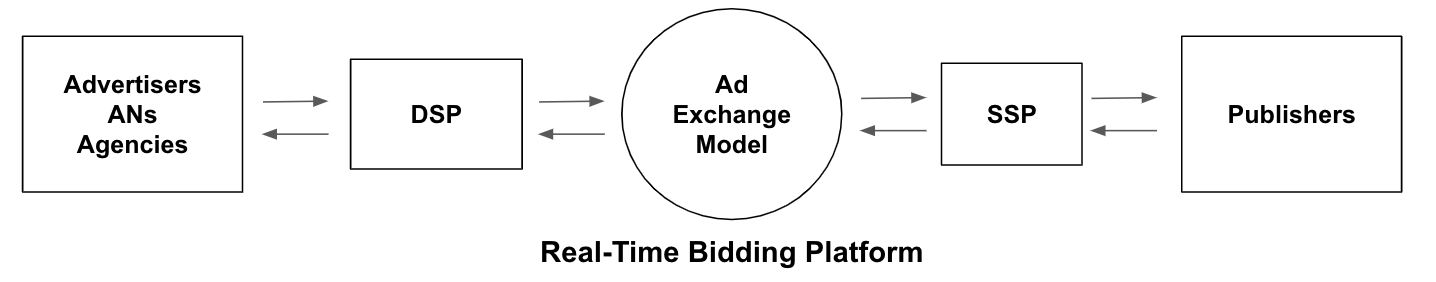}
\caption{Structure of the most important modules and roles in a Real-time Bidding platform.}
\label{fig:Methodology}
\end{figure}

In online advertising, there exist a conflict of interests between advertisers and publishers, and a trade-off between the two competing interests is needed. On one end, advertisers desire low prices for their impressions and on the other, publishers demand higher income for displaying adverts on their sites \cite{yuan2012sequential}. As shown in figure \ref{fig:Methodology}, there are two platforms that define the RTB ecosystem: the demand-side platform (DSP), representing the interests of advertisers, and the supply-side platform (SSP), representing the interests of publishers. The DSP, manages advertisers and ANs campaigns efficiently, bidding directly on the auctions. On the other hand, the SSP aims at managing and optimising publishers' web spaces. The SSP distributes the information of a publisher across multiple platforms whenever a user visits a website and selects the most cost-effective advert \cite{zhang2014optimal}. In RTB, the publisher can have a reserve price, that is to say, a minimum price to sell the impression. Yuan et al. \cite{yuan2014empirical} have addressed the issue of how to efficiently calculate the optimal value for the reserved price.

Performance-based display advertising in RTB platforms usually implements the CPC payment model (Cost-per-click), in which the advertiser pays only when adverts generate a click or the Cost-per-acquisition (CPA), in which advertisers pay when a conversion derived from the displayed adverts takes place \cite{chen2011real}. Thus, in order to select the best advert, the DSP implements ML models to estimate the probability of generating a conversion or a click from a given advert \cite{lee2012estimating,miralles2017methodology}.

\subsection{Real-Time bidding campaign optimization techniques}

RTB campaign optimization differs from search engine optimization in the way that an advertiser does not bid on a particular keyword, but instead, bids for individual impressions \cite{wang2017display}. To optimise an RTB campaign, several factors are analysed such as the duration of the campaign, the preferences of each advertiser segment, the competitors' bids and strategies, the reserve price of the publishers, or the number of networks \cite{yuan2014survey}.

Predicting the optimal bid price for each impression is one of the most recurring challenges in  RTB campaigns. When the price is too low, it is likely that it will not win the bid, on the contrary, if the price is too high, it will be paid in excess. To optimise the price paid for a campaign, an effective bidding strategy is essential. The bidding strategy determines the optimal quantity an advertiser pays for each impression. The effectiveness of the optimization strategy depends on different parameters, such as the payment method (advertisers can pay for impressions, clicks or actions), whether the bid is transparent or not, or the bidding mechanism \cite{yuan2014empirical}.

There are mainly two approaches to estimate the optimal price. The first is based on game theory, where both advertisers and publishers make intelligent decisions based on others behaviour \cite{roth1992two}. The game theory approach called Bayes-Nash equilibrium is a popular choice \cite{gomes2014bayes}. The second approach is based on statistics, through the observation of the behaviour of the clicks, bids and conversions, it can be appreciated that sometimes periodic patterns are presented. That fact suggests that models that consider the time as a variable will perform better when calculating the optimal price \cite{yuan2013real}.

Ghosh et al. \cite{ghosh2009adaptive} introduced a different perspective that includes two scenarios for estimating the bidding price: one in which the bidding mechanism is transparent and all participants are informed of the winning price of the bid, and another one in which only the advertiser who has won the bid is informed of the winning price. Perlich et al. \cite{perlich2012bid} have examined the problem of setting the correct price according to the type of user, the type of message to be sent, and the specific moment based on supervised models and price auction theory. The process is as follows: first, the value of a particular advert is estimated, then, a bidding strategy based on the estimated price is applied. Additionally, it proposes a more aggressive bidding strategy based on a step function to set the price of bids.

Traditionally, the RTB ecosystem made use of the Generalised Second-Price (GSP) auction mechanism. In GSP, the winner (the highest bid), instead of paying the amount of the bid, the advertiser pays the amount of the second-highest bid. However, the company Getintent, after analysing 338B impressions, reported that since 2017, many RTB companies decided to change their auction model to First-price auction, growing from 5.8\% in December 2017 to 43.3\% in March 2018 \cite{First-price-auctions}. The main reason for this change is that First-price auctions are more transparent by definition (the winner pays exactly what they bid for), and advertisers are more confident about the price they are changed.

In RTB most advertisers apply strategies to maximise the utility to select the most appropriate set of users for their campaigns by minimising the cost. Usually, it is preferable to lose a bid than to pay an excessive price for an advert.  In RTB, a dynamic approach is more appropriate for estimating the amount to bid than a static approach, because the online advertising markets are very dynamic (there are permanently new advertisers, publishers, bidding strategies, and so on) \cite{borgs2007dynamics}. To define a dynamic strategy, the supervised method must be updated frequently \cite{lang2012handling}. Additionally, advertisers prefer their adverts to be displayed uniformly over time to reach a more diverse audience \cite{lee2013real}. To this end, two strategies can be applied: modifying the value of the bids, or randomly selecting whether or not to participate in a bid \cite{xu2015smart}.

It is worthwhile considering the approach based on the emerging machine learning branch called reinforcement learning proposed by Busoniu et al \cite{busoniu2008comprehensive}. In this new technique, the system learns from its own experience through a reward function. The reward is positive when the system performs well and negative when the system fails \cite{busoniu2008comprehensive}. In the long term, the system favours those actions that maximise the number of rewards over several trials \cite{sutton1998introduction}.

Reinforcement learning is a popular choice dealing with decision making that has been recently applied in many domains such as robotics, bots playing chess, and also RTB. Cai, H. et al. \cite{cai2017real} presents a paper in which the bidding strategy is formulated as a reinforcement learning problem. In the Markov decision process, the agent represents the advertiser, the market is the environment, the parameters of the campaign (lifetime, budget ...) are the current state, and the bids of the agent are the actions. The presented model is called Reinforcement Learning to Bid (RLB) and it surpasses the performance of other state-of-art methods.

\subsection{\textbf{Optimizing Real-Time bidding campaigns through parameter configuration}}

Internet publicity platforms provide monitoring tools that allows advertisers to access useful information such as if the user clicked on the advert, the amount of time the user spent on the page, which users have visited the website before, and which items have been bought \cite{yan2009much}. With the help of these tools, the performance of an online campaign can be evaluated very accurately just a few minutes after a campaign is launched. With this advantage, the campaign can be adjusted continuously to maximise the profitability by altering the configurations which impeded the profitability.

One of the main benefits of online advertising is that it enables advertisers to segment their target audience in a precise way which in traditional media is much more difficult to implement \cite{goldfarb2011online}. In online publicity, an advert is shown only to the specific set of users selected by some criteria whereas, in television and radio, adverts are shown or voiced to all users regardless of their interests and characteristics.

Furthermore, online publicity networks identify users' profiles by exploiting features such as browsing history, age, gender, operating system, or device type. This strategy of directing campaigns to a small group of people having common interests is useful and is known as microtargeting \cite{goldfarb2011online,miralles2018novel}

Improving the overall performance of the campaign can be addressed through an appropriate configuration of parameters related to users and websites. Y. Liu et al. \cite{liu2012finding} showed that by using user features along with metadata from web pages, it is possible to estimate which users are more likely to generate a conversion. Additionally, it is also possible to create models to predict the probability of new users accessing to new web pages \cite{liu2012finding}.

Our contribution approach is related to that of Y. Liu et al. \cite{liu2012finding} which aims at optimising RTB campaigns through the selection of attributes from the users and web pages. Our approach is focused on finding parameter configurations that maximise both the number of visits and the average conversion probability for those visits. Additionally, the parameter configuration technique is combined with other approaches such as multiple configurations, cost and profitability thresholds, configuration extrapolation, and increasing the search space. The combination of multiple approaches with the configuration optimization can potentially increase even more the obtained performance.

\section{RTB Campaign Optimization methodology} \label{sec:methodology}

In this section, we present the proposed methodology to optimise RTB campaigns based on attribute configuration instead of doing so by applying the most popular approach in the literature, that is, developing a better bidding strategy.

\subsection{Description of the RTB Campaign Optimization methodology}\label{sec:methodology-description}

\begin{figure}
\centering
\includegraphics[scale=0.5]{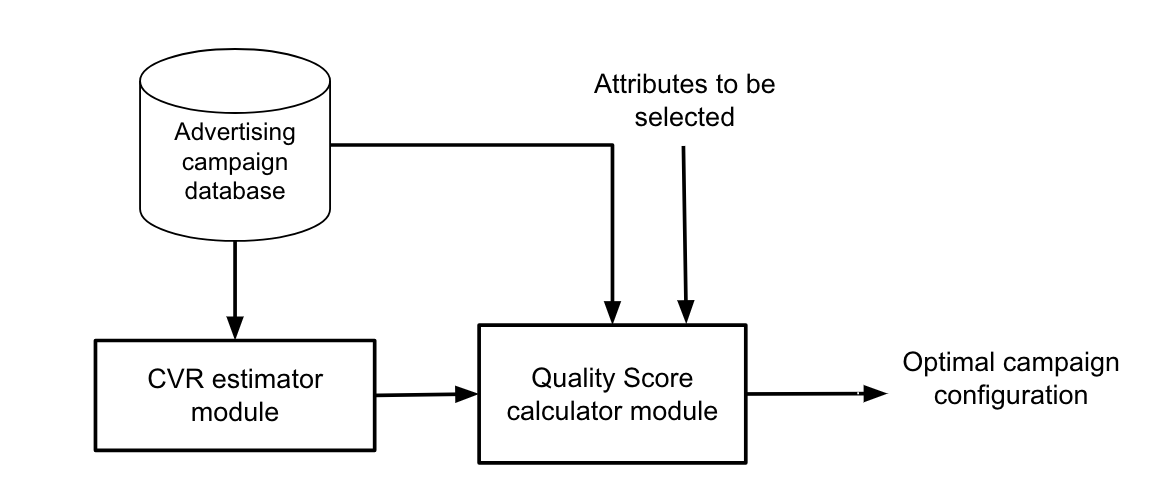}
\caption{The methodology used to find out the bests campaign configuration.}
\label{fig:Methodology1}
\end{figure}

Our approach is based on detecting configurations (sets of parameters from users and web attributes, along with their values) to suggests them to advertisers so they can display adverts more effectively. An example of parameter configuration could be: \{Browser = "Chrome", Device OS = "Android",  Age = "25-34", Gender = "Male", Time = "10:00-11:00", Country = "UK"\}. Figure \ref{fig:Methodology1} shows the overall architecture of the proposed method which is composed of two modules: the $Conversion\,rate$ (CVR) estimator and the $Quality\,Score$ calculator (QSC).

Our approach begins with the CVR estimator module that calculates the conversion probability (the probability that a conversion will be generated when an advert is displayed to a user) of each visit based on historical data. The CVR module calculates the probability of a conversion from a given impression and it is a supervised model based on logistic regression (it is briefly described in section \ref{sec:CVR-model}). The second module is the QSC which scores all of the possible configurations (i.e., combinations of different attributes). The QSC module makes use of all the attributes of the impressions and optimises them to achieve the highest performance. Firstly, QSC explores all the possible configurations with a single attribute, then, it continues with two attributes, and so on and so forth until it finally reaches all the combinations with all the attributes. All configurations are ranked according to $Quality\,Score$ metric defined by equation \ref{eq:Quality-score}.

To build the CVR estimator module, we used a dataset from Criteo \cite{diemert2017attribution}. The dataset is ordered by time and represents 30 days. See section\ref{subsec:dataset-desc} for more details on the dataset. We trained the supervised model with the first 6.46 M rows of the dataset and then, we calculated the conversion probability over the rest of the dataset, which we considered the testing set and has 10M rows. For evaluating the performance of our methodology, we included the prediction of the trained model as a new column of the testing set, and we stored it in the advertising campaign dataset module.

The metric called $Quality\,Score$ ranks how good a configuration is. The $Quality\,Score$ favours configurations with high profitability as long as they have the minimum number of visits required by the advertiser. The $Quality\,Score$ is calculated for all the possible configurations without repetition (discussed in detail in the following Section \ref{subsec:rank-configuration}). Equation \ref{eq:combinations} shows the formula to calculate the number of combinations. Finally, the higher-ranked configurations scored by the QSC are offered to advertisers to increase the performance of their campaigns.

\begin{equation}
Combinations \, without \, repetitions = \sum_{k=1}^{n} \frac{n!}{k!\left(n-k\right)!}
\label{eq:combinations}
\end{equation}

\subsection{Design of the Conversion Rate prediction model} \label{sec:CVR-model}

In this section, we discuss the design choices that were made to construct the CVR estimator model. To estimate the CVR, a database of displayed adverts is used, where each row is the displayed advert on the website of a publisher. The dataset contains attributes related to the user, the publisher's web page, the campaign identifier, the indicator of whether the conversion was generated, and the price paid for an impression.

Equation \ref{eq:profitab} shows the formulation of the profitability of an impression using the $CVR$ and the $Impression\,_{Price}$.  The $CVR$ is the probability that a conversion is generated from an advert\footnote{We used the supervised model to estimate the $CVR$ value proposed in \cite{lee2012estimating,ahmed2014scalable}}, and the $Impression\,_{Price}$ is the price of the impression whose value is extracted from the dataset.

According to equation \ref{eq:profitab}, the profitability will be higher for those impressions that have a high probability of generating a conversion (numerator) and a low price (denominator).

\begin{equation}
Impression\,_{Profitability} = \frac{CVR}{Impression\,_{Price}}
\label{eq:profitab}
\end{equation}

We used a train/test split to build and validate the performance of the model. Figure \ref{fig:rowselection} shows a snapshot of the typical dataset that the proposed model makes use of. The dataset has a total of ten columns, out of which nine are categorical (cat1, ..., cat9) which forms the predictor variables for the model $e_i, i = 1,...,9$, and the final column is the $Profitability$  which is the target or the output variable of the model $y$. The output, $y$, represents the probability that a conversion (a purchase, a call, the filling of a form...) takes place. For instance, a predicted value of 0.225 means that the model estimates that the user has 22.5\% chance of generating a conversion from the displayed advert. The model is used to expressed $p$, where $p = P(y|e_i, \forall i = 1,...,9)$.

The CVR model was built using a well-known methodology for making predictions based on a Logistic Regression model \cite{chapelle2015simple,li2011hashing}. We configured the adaptive learning rate value by following the criteria defined in \cite{mcmahan2013ad}. Such paper explains that the value of alpha highly depends on the nature of the dataset and the type of data. In our particular case, we used several parameters and choose the one that gave the best performance. The results and performance of the logistic regression algorithm are presented in section \ref{sec:model}.

To create the model, we used a simple but effective technique called hashing trick. The hashing trick is used to reduce the sparsity of the values of the dataset, not the number of features in the dataset\cite{li2011hashing}. This technique allows to generate precise models and yet consuming little memory and computational resources. The hashing trick is recommended when the dataset is large, it consists of applying the modulus operation to each of the elements of the dataset $e_{i}\,mod\,D\,=\,R$, where $e_{i}$ is an entry, $D$ is a constant whose value should be high enough to avoid collisions, and $R$ is the remainder of the Euclidean division (the hashed value).

Collisions occur when different elements produce the same number after the application of the modulus operation. This trick works well with Logistic Regression models because it makes use of two vectors. The first vector, $n$, is used to store integers and it counts the number of times a value is repeated when the hash function is applied. The second vector, $ w $, stores real numbers that represent the weight associated with each value of the dataset. The greater the value of a weight, the greater the value of the output will be \cite{li2011hashing}. 

The original dataset can have $N$ different values for each category, where $N$ can be bigger than the length of the arrays $n$ and $w$. The hashing trick divides the values of $N$ by $D$, where $D$ = length($w$) =  length($n$), so the sparsity of the values will be at maximum $D$. This trick works very well because it reduces the amount of required space of memory from a computer to $M$, and despite being a simple trick, the results report that it has a very high performance in performance, time and consumes low memory \cite{li2011hashing}.

Initially, all of the elements of both vectors are set to zero, afterwards, the elements of $n$ and $w$ are updated using equations \ref{eq:frequency} and \ref{eq:weights}.

\begin{equation}
n[i]=n[i]+1
\label{eq:frequency}
\end{equation}\begin{equation}
w[i]=w[i]-\frac{\alpha \times (p-y)}{\sqrt{n[i]+1}}
\label{eq:weights}
\end{equation}

Once the vectors \textit{w} and \textit{n} have been updated, we make the prediction of the CVR for each impression using equation \ref{eq:sigmoid}, which is the output of the model calculated as a sigmoid function. 

\begin{equation}
p=\frac{1}{1+exp\left(-{\displaystyle \sum_{i=1}^{N}}w[f_{i}]\right)}
\label{eq:sigmoid}
\end{equation}

Where each variable is described as follows:
\begin{itemize}
\item $i$ represents the index of the arrays $w$ and $n$.
\item $w[i]$ represents the weight of $i-th$ element.
\item $n[i]$ represents the number of times that the value of the $i$ appears after the application of the hashing trick to the elements of the original vector.
\item $p \in [0,1]$ represents if there was a conversion and it is also the estimated output value.
\item $\alpha$ represents the heuristic adaptive learning rate which is used to optimise the weights of the model. The higher the value of the learning rate, the faster it adapts but it may overshoot the gradient. By contrast, the lower the value, the more time it takes to coverage.
\end{itemize}

\subsection{Design of the algorithm to rank configurations of the campaign} \label{subsec:rank-configuration}

In this phase, we attempt to find optimal configurations that maximise both, the number of visits fitting the configuration and the average profitability of those visits. To meet the objective, we defined a criterion called $Quality\,Score$ as shown in equation \ref{eq:Quality-score}. The $Quality\,Score$ rewards configurations that presumably have high profitability (sets of visits with a high probability of conversion at a low price) and also, ensuring a sufficient number of impressions for the advertiser campaign.

\begin{figure}
\centering
\includegraphics[scale=0.5]{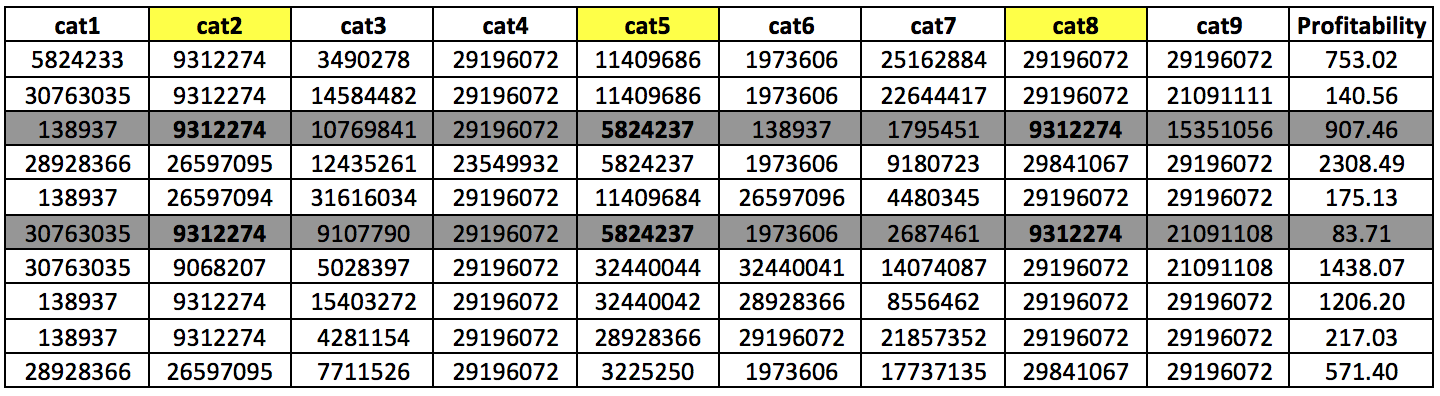}
\caption{Extracting from the original dataset those visits that fit the configuration: (columns 2, 5, and 8 with values 9312274, 582437 and 9312274 respectively.}
\label{fig:rowselection}
\end{figure}

To begin calculating the value of a configuration, we first estimate the average profitability. To this end, we score the profitability (using equation \ref{eq:profitab}) for all the impressions that fit the configuration, and then, we calculate the average. Even though the configurations may have some unprofitable impressions, these impressions are offset by the rest of the impressions making the metric robust. Furthermore, we multiply the $Profitability\,_{Average}$ by the minimum between the number of impressions that satisfy the configuration and the number of visits required by the advertiser, i.e., $\min(rows(D'),T)$ as shown equation \ref{eq:Quality-score}. The $T$ represents a threshold, and it is used to avoid converging to configurations with an excessive number of visits. We should bear in mind that we are looking for configurations with a sufficient number to cover the number of visits required by the advertiser.
\begin{equation}
Quality\,Score = Profitability\,_{Average} \times \min(rows(D'),T)
\label{eq:Quality-score}
\end{equation}
In the following lines, it is given an example of how the process of calculating the $Quality\,Scores$ for a combination of attributes works. Let's imagine that we are testing the combination of columns (2, 5, 8). First, we will select the second, fifth and eighth attribute from the dataset. The result will be a subset with three selected attributes from the original dataset. From this subset, we select the unique records (i.e., rows) and calculate the $Quality\,Scores$ for each record. As shown in Figure \ref{fig:rowselection}, the first unique value of the subset are values: 9312274, 582437 and 9312274 respectively. Then, we operate with the subset called D' (where the second, the fifth and the eighth attribute have values 9312274, 582437 and 9312274 respectively) are extracted from the original dataset. For this subset D', where D' $\subseteq$ D, the $Quality\,Score$ is calculated using equation \ref{eq:Quality-score}.

To speed up the algorithm, we created a list of all the configurations that do not have enough visits. Then, we add the following condition: if a configuration of a campaign is a subset of a previously discarded configuration, then this new configuration gets dropped as well. Such action was taken on the basis that a subset cannot be greater than the set that contains it. For example, for attributes (1, 3, 4, 7) with values (458, 47, 58, 58) respectively, we check if the list of configurations with not enough visits contains any combination without repetition of these attributes with precisely these same values. To do so, we first check in the list, combinations of one element (attribute 1 with value 458, attribute 3 with value 47, and so on). Later, we look for in the list for combinations of two elements (attribute 1 with value 458 and attribute 3 with value 47, attribute 3 with value 47, and attribute 4 with value 58, and so on), and the same with the combinations of 3 elements. This simple improvement significantly reduces the running time to find out the optimal solution. This small improvement results in completing an experiment (see experiment II in the later section) within 193.82 seconds compared to 75 hours without applying this trick. 

\section{Experiments}

In this section, experiments are carried out to evaluate the yield of the proposed methodology. First, we describe the preprocessing step applied to the dataset before running the experiments. Secondly, we explain in detail how to build the model to estimate the \textit{CVR} value: we evaluated the built models using the well-known metrics and predicted the \textit{CVR} of the RTB impressions. Next, we explain the algorithm used to find the optimal configuration. Finally, we discuss the five carried out experiments, and in each experiment, we demonstrate different approaches in the context of attribute selection.

The proposed experiments have common steps which are presented in the algorithm \ref{alg:basic-steps}. The steps "Calculate the Quality Score" and "Build the CVR estimation model" have been explained in detail in the previous section (see section \ref{sec:methodology}).

\begin{algorithm}
\footnotesize{
\begin{algorithmic}[1]
\State Build the CVR estimation model 
\State Predict the CVR for all impressions
\State Generate all possible campaign configurations
\ForAll {Unique values of a combination}
\State Calculate the Quality Score
\State Store solution in the dataset
\EndFor
\State Return the best solution from the dataset
\end{algorithmic}}
\caption{ Selecting the best online campaign configuration}
\label{alg:basic-steps}
\end{algorithm}

To carry out the experiments, we used Python 3.6.1 and Anaconda runtime (x86\_64). We performed all experiments using MacBook computer having macOS High Sierra (2.3 GHz Intel Core i5, 8GB 2133 MHz, L2 Cache:256kb, L3: 4MB).

\subsection{Dataset description} \label{subsec:dataset-desc}

We conducted the experiments using a dataset from the Criteo company \cite{diemert2017attribution}. The dataset consists of real data collected from internet traffic for 30 days. The dataset is a subset of the total visits from that period. Each row from the dataset contains information about an advert displayed on the website of a publisher. The dataset is anonymised for privacy concerns.
\\
The dataset is composed of the following fields:
\begin{itemize}
\item Timestamp: it starts at 0 and sorts rows according to the displayed time.
\item UID: Unique code to identify the user.
\item Campaign: unique code to identify the campaign.
\item Conversion: it uses "1" if a conversion in the following 30 days takes place. In other cases, "0" is used.
\item Conversion\_timestamp: the exact moment where the conversion derived from the impression took place. "-1" is used when there is no conversion.
\item Conversion\_id: the code of the conversion derived from the impression. It enables the creation of the timeliness (in case they are required).
\item Attribution: binary field to indicate if the conversion was connected to Criteo ("1" means affirmative).
\item Click: binary field to indicate if the user clicked the impression ("1" is affirmative)
\item Click\_pos: when there are several clicks from a user before a conversion, this value represents the position of the click in the current impression.
\item Click\_nb: number of clicks before the conversion (the same user can click several times in adverts of the same campaign).
\item Cost: a transformed version of the price paid by Criteo.
\item Cost-per-order(CPO): the amount paid by the advertiser when the conversion is assigned to Criteo.
\item Time\_since\_last\_click: indicates in seconds the amount of time elapsed from the last click.
\item Cat[1-9]: these nine fields represent features related to the user and the publisher. These fields are encrypted and had been hashed by applying the well-known hashing trick to reduce the dimensions \cite{li2011hashing}. In addition, these values are used to create a supervised model for predicting the conversion probability.
\end{itemize}

\begin{figure}
\centering
  \includegraphics[scale=0.6]{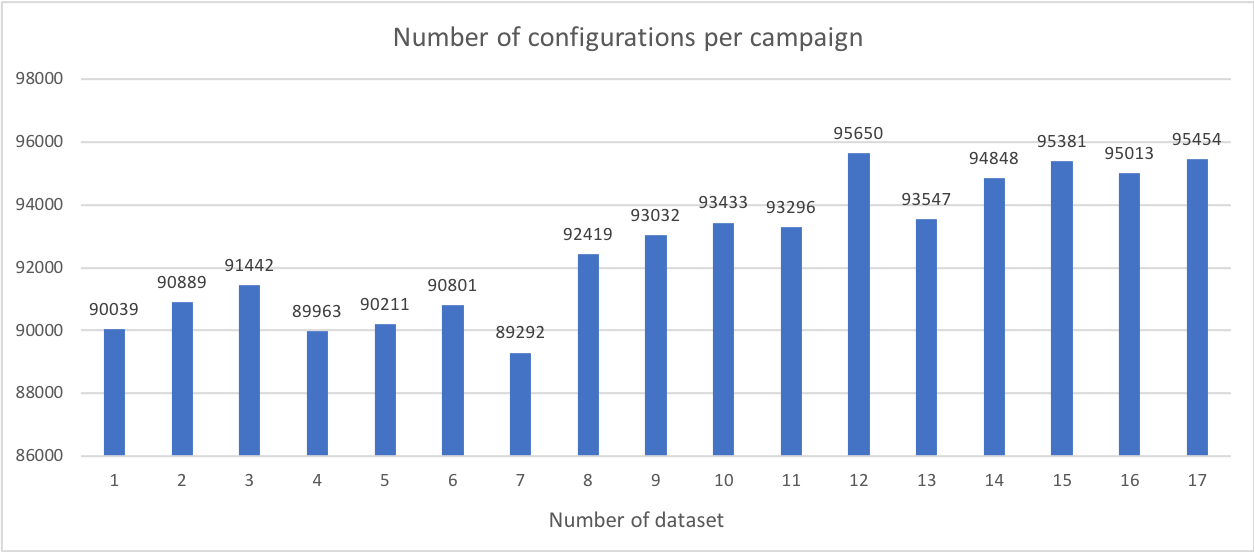}
  \caption{As we can see in the graphic there are a lot of combinations in each of the 17 datasets.}
  \label{fig:configurations}
\end{figure}

\begin{figure}
\centering
  \includegraphics[scale=0.6]{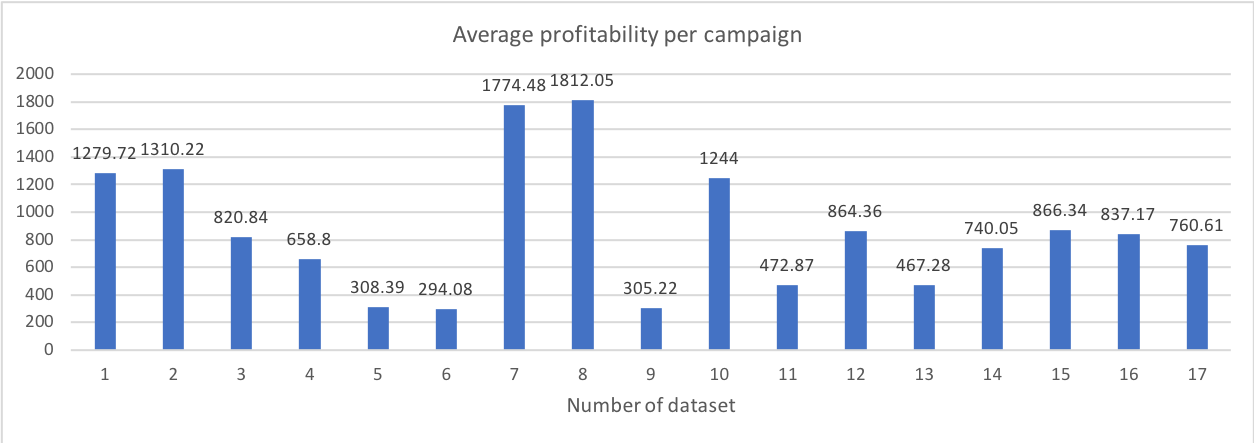}
  \caption{Average profitability for each of the 17 advertising campaigns.}
  \label{fig:avg_performance}
\end{figure}

The original dataset has a total of 16,468,027 instances from different campaigns. First, we split the dataset into the training set (first 6,468,027 rows) and the testing set (last 10 million rows). After the split, we built the CVR model using the training set, and we predicted the probability of generating a conversion for the instances of the testing set. In the testing set, there are four campaign identifiers with more than 200 thousand visits (the identifiers of the campaigns are: 10341182, 15398570, 17686799, 30801593). We further divided each of the four campaigns into two parts of 100 thousand visits each. There are also nine other campaign identifiers with more than 100 thousand visits: 5061834, 15184511, 29427842, 18975823, 6686701, 28351001, 26852339, 31772643, 497593. 
We discarded the rows above 100 thousand so that all the datasets had the same dimensions. Finally, we got a total of 17 datasets of 100 thousand visits each (8 as a result of dividing each of the 4 campaigns with 200k into two campaigns of 100k and the other nine campaigns with 100k visits). All the experiments were carried out using the 17 datasets and figure \ref{fig:configurations} shows the total number of possible configurations in each dataset. Figure \ref{fig:avg_performance} shows the Average profitability of each campaign.

\subsection{Implementation of the Conversion Rate model} \label{sec:model}

As described in section \ref{sec:methodology}, first we built and trained the predictive CVR model. In addition to that, we converted all features from string values to integer values and applied modulus $2^{20}$ to calculate the hashing table. Later, we applied the logistic regression method to build the prediction conversion model with a \textit{learning rate} $\alpha = 0.1$ and $D = 2^{20}$ as the length of arrays $n$ and $w$. Algorithm \ref{alg:Algorithm-to-Conversion-Model} summarises the implementation of the Conversion Rate model for both training and testing it.

The results of the CVR model over the testing set are shown in table \ref{tab:conf_matrix}. In table \ref{tab:PredictionMetrics}, we can see that the Accuracy is 95.10\% and the Area Under the Curve (AUC) is 82.99\%, while the Logarithmic loss (a metric commonly used in this type of problem) has a value of 0.1572, which is quite accurate.

\begin{table}[htbp]
\caption{Conversion probability confusion matrix.}
\centering
\begin{tabular}{l|l|c|c|c}
\multicolumn{2}{c}{}&\multicolumn{2}{c}{True values}&\\
\cline{3-4}
\multicolumn{2}{c|}{}&Positive&Negative&\multicolumn{1}{c}{Total}\\
\cline{2-4}\textbf{}
\multirow{2}{*}{Predicted values}& Positive & $9490262$ & $14271$ & $9504533$\\
\cline{2-4}
& Negative & $475589$ & $19878$ & $495467$\\
\cline{2-4}
\multicolumn{1}{c}{} & \multicolumn{1}{c}{Total} & \multicolumn{1}{c}{$9965851$} & \multicolumn{    1}{c}{$34149$} & \multicolumn{1}{c}{$10000000$}\\
\end{tabular}
\label{tab:conf_matrix}
\end{table}

\begin{table}[htbp]
\caption{Metrics of the Conversion probability model.}
\centering { \noindent\adjustbox{max width=\textwidth}{%
\begin{tabular}{|c|c|c|c|c|c|c|c|c|c|c|}
\hline
Log Reg &  MAE &  MSE &  RMSE &  AUC & Acc & Avg Acc & Sens & Spec & Prec & F-1 \\ \hline
0.1572 & 0.0807 & 0.0411 & 0.2027 & 0.8299 & 0.9510 & 0.5193 & 0.0401 & 0.9985 & 0.5821 & 0.0750 \\ \hline
\end{tabular}}}
\label{tab:PredictionMetrics}
\end{table}
\begin{algorithm}
\footnotesize{
\begin{algorithmic}[1] 
\Require Dataset 
\Ensure  Prediction Conversion model
\vspace{0.25cm}
\State $data \gets$ Randomly select 12M samples visits from the original dataset 
\State $data_{hash} \gets $ Apply the hashing trick to $data$
\State Divide $data_{hash}$ into $training$ and $testing$ sets \Comment 6.3M $training$ and 10M for $testing$
\State {$ D  \gets 2^{20}$} \Comment Length of $w$ and $n$
\State {$ \alpha  \gets 0.1 $} 
\State {$ w  \gets  \begin{bmatrix} 0 & 0 & 0 & \cdots & 0 \end{bmatrix}$} of length $D$
\State {$ n  \gets  \begin{bmatrix} 0 & 0 & 0 & \cdots & 0 \end{bmatrix}$}  of length $D$
\vspace{0.25cm} \\ $\triangleright$ Training the CTR model 
\ForAll{$row_{k}\in training$}  \Comment $row$ represents a user visit in which adverts display their adverts	
\State {$ s  \gets 0 $} 
\ForAll{$f_{i}\in row_{k}$}    
\State {$s \gets s + w[f_{i}+1]$}
\EndFor
\State $p_k \gets 1 /(1+exp(s))$ 

\ForAll{$f_{i}\in v_{k}$}    
\State {$w[f_{i}] \gets w[f_{i}] - \alpha(p_k - y_k)  / (\sqrt{n[f_{i}] + 1})$} 
\State {$n[f_{i}] \gets n[f_{i}]+1$}
\EndFor
\EndFor  
\vspace{0.25cm} \\ $\triangleright$ Testing the Conversion model 
\ForAll{$row_{k}\in testing$} 	\Comment $v_k$ is a $testing$ sample	
\State {$ s  \gets 0 $} 
\ForAll{$f_{i}\in row_{k}$}    
\State {$s \gets s + w[f_{i}+1]$}
\EndFor
\State $p_k \gets 1/(1+exp(s))$
\EndFor
\State Compute the accuracy of the model  
\end{algorithmic}
}
\caption{Training and testing the prediction Conversion model.}
\label{alg:Algorithm-to-Conversion-Model}
\end{algorithm}

\subsection{Implementation of the algorithm to find the best solution}

To select the optimal configuration given the requirements of the advertisers, we executed algorithm \ref{alg:gaBestComb} which in turn calls algorithm \ref{alg:CalculateSolution}. The idea is to test all possible combinations of attributes and select the unique values for each combination. For example, if the attributes (1, 5) have the values (85, 58), (7714, 424), (596, 3458), (85, 58). Then, the last one will be rejected because it is the same as the first one. For each unique value, we choose from the dataset all the rows with the value 85 in attribute 1 and the value 58 in attribute 5. For that subset of the dataset, we select the average profitability and the number of rows from the dataset that match with that configuration. The idea is to explore all the possible combinations and store the ones with higher values.

\begin{algorithm}
\footnotesize{
\begin{algorithmic}[1]
\State Comb $\gets$ List of all possible column combinations \Comment (1),(2),(3),...(1,2),(1,3),(1,4),...(1,2,3),(1,2,4),...	
\State List $\gets$ \{ \} 
\ForAll{$d_{i}\in Datasets$} \Comment There are 17 datasets
\State Limit $\gets$ 5,000
\While{(Limit $<=$ 30,000)}
\State  $Solution \gets Calculate\_best\_solution(d_{i}, Limit, Comb)$ 
\State List $\gets$ List + Solution 
\State Limit $\gets$ Limit + 5,000
\EndWhile
\EndFor
\end{algorithmic}}
\caption{Detecting the best combination of the dataset.}
\label{alg:gaBestComb}
\end{algorithm}

\begin{algorithm}
\footnotesize{
\begin{algorithmic}[1]
\Require Data, Comb, Limit
\Ensure  Tuple
\State RejectSolution $\gets$ \{ \} 
\State SolList $\gets$ \{ \} 
\ForAll{$c_{i}\in $ Comb} \Comment Select only columns from combination of the Data
\State Data1 $\gets$ Subset(Data[,$c_{i}$]) \Comment The value of Data1 is a subset deep copy of Data
\State Data1 $\gets$ Unique (Data1) \Comment If a configuration is repeated we do not have to calculate the value again
\ForAll{row $\in$ Data1}
\For{i $\in$ 1:Length(row)}
\State Data2 $\gets$ Subset(Data1[,Comb[i]] = row[i])
\If {! RejectSolution.exists($c_{i}$, row)} 
\If {Rows(Data2) >= Limit} 
\State $Profit_{Avg} \gets$ mean(Data2[,Prof])
\State Size $\gets$ Rows(Data2)
\State Fitness $\gets Profit_{Avg} \times$ min(Size, Limit) 
\State Sol $\gets$ Tuple({Fitness, $Profit_{Avg}$, Size, $c_{i}$, row})
\State SolList $\gets$ SolList + Sol
\Else
\State  RejectSolution.append(Sol[1]) \Comment Add rejected solution to the list
\EndIf
\EndIf
\EndFor
\EndFor
\EndFor
\State  BestSol $\gets$ OrderbyFitness(SolList) \Comment Order by Fitness
\State \Return BestSol
\end{algorithmic}}
\caption{Calculating the best solution.}
\label{alg:CalculateSolution}
\end{algorithm}

\subsection{Description of the experiments}
As shown in table \ref{tab:Solutions}, for each of the configurations, we collect the following information:
\begin{itemize}
\item Average profitability: indicates the average profitability of the selected impressions in the configuration campaign.
\item Dataset size: indicates the number of rows of the dataset. In experiment II a number of rows are removed each time a configuration is selected.
\item Nº Rows: indicates the number of rows of the best configuration, which is the configuration with the highest value.
\item Selected columns: indicates the selected attributes by the best configuration.
\item Time (Sec): required time to find the best configuration.
\item Values for selected columns: in addition to which columns/attributes have to be selected, we are also interested in the values that optimise those parameters.
\end{itemize}
\begin{itemize}
\item Quality Score: as indicated previously, this metric is used to show how good a configuration is, given the number of visits required by the user.
\item Dataset size: the remaining size of the dataset after subtracting the selected configurations.
\end{itemize}

\begin{figure}
\centering
  \includegraphics[scale=0.6]{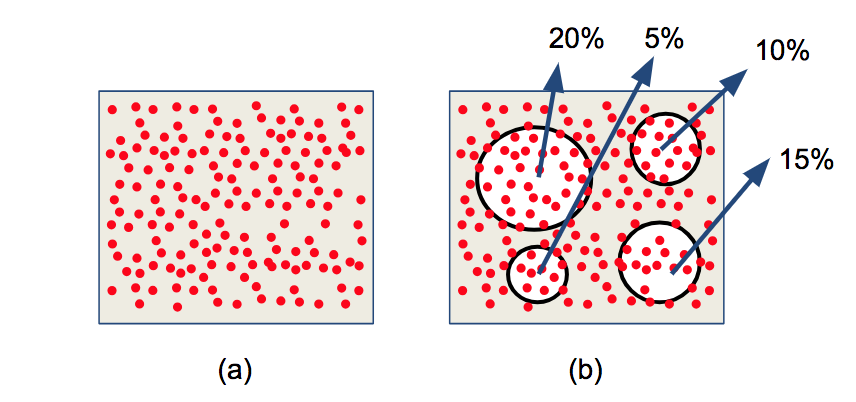}
  \caption{Each configuration has two parameters: the average profitability and the number of visits matching the parameter configuration. The first picture (a) shows all the visits and the second (b) the required visits by the advertisers.}
  \label{fig:ExplanationI}
\end{figure}

Next, we make a small description of the carried out experiments:
• \textbf{Experiment I}: In the first experiment, we intend to improve the campaign performance by optimising the parameter configuration. Generally, the more specific the campaigns are, the better the yield is, but, on the other hand, the number of visits that match the configuration is lower. As shown in figure \ref{fig:ExplanationI}, the original datasets have 100K samples. In this experiment, we select the best configuration from the dataset with at least 5k (5\% of 100k) visits.  Then, we repeated the same process increasing the limit of selected visits by the factor of 5k until reaching 50k. If advertisers require a small number of visits, the profitability will be very high, but, to perform a successful campaign it is generally required to display a sufficient number of adverts. The exciting aspect of this experiment is to visualise how profitability decreases as the number of demanded visits increases.

\begin{figure}
\centering
  \includegraphics[scale=0.6]{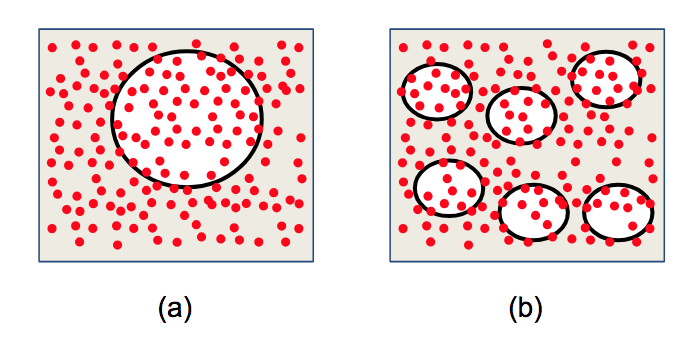}
  \caption{Selecting a large group as in (a) is computationally less expensive than selecting small groups as in (b), but the summation of small groups gives better results.}
  \label{fig:ExplanationII}
\end{figure}

• \textbf{Experiment II}:  The motivation of the second experiment is to answer if it is better to optimise $N$ campaigns with a small number of visits $S$, or to optimise a campaign with a big number of visits $B$, where the summation of the $S$ visits of the $N$ small campaigns is similar to that of $B$. In figure \ref{fig:ExplanationII} the two approaches can be seen graphically. Optimising more campaigns has a higher computational cost, but it is possible that a better solution offsets these costs. To avoid the problem of the intersection of visits, which occurs when two or more campaigns share visits, we perform the experiments sequentially so that when the first configuration campaign is selected, all the visits that match with such configuration are removed from the dataset. The same process is repeated until all the campaigns are chosen. This experiment is performed for the following number of visits required by the advertiser: 5k, 10k, 15k, 20k, 25k and 30k. Finally, we compare the performance of selecting a campaign with these configurations with picking configurations with 1k and 2.5k visits respectively. For example, we compare a configuration that matches at least 5k visits with the summation of two configurations with at least 2.5k, and with five configurations with at least 1k visits.

\begin{figure}
\centering
  \includegraphics[scale=0.6]{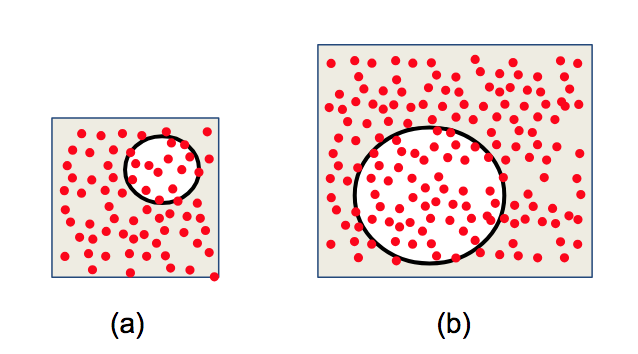}
  \caption{ If it is possible to extrapolate a good solution from a slice of the campaign (a) instead of the using the whole dataset (b), then profitable configurations can be obtained at a lower computational costs.}
  \label{fig:ExplanationIII}
\end{figure}

• \textbf{Experiment III}: In this experiment, we investigate the effect of finding a profitable configuration from a subset of the dataset and observe if the same configuration stills performing well over the entire dataset. If the finding is positive, it will have a substantial advantage in terms of computational costs (time and memory). Figure \ref{fig:ExplanationIII} shows the impact of configurations extracted from the subset and extrapolated over the whole dataset. In this experiment, we extrapolate from 10\%, 20\%, 30\%, and so on until we reach 100 \% of the dataset.

\begin{figure}
\centering
  \includegraphics[scale=0.6]{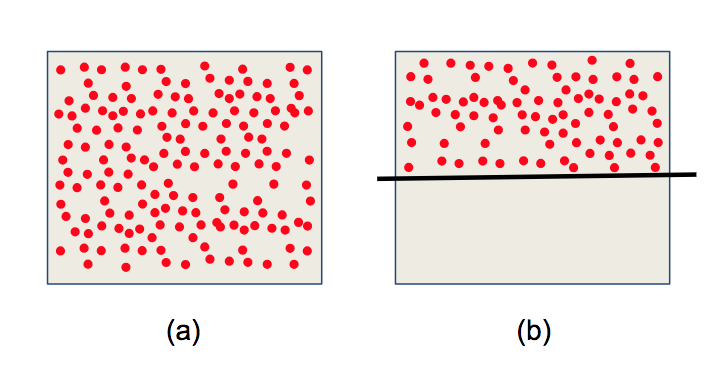}
  \caption{Discarding visits over a certain price limit or below a certain profitability value as in (b) can boost the results than when the whole campaigns (a) are used.}
  \label{fig:ExplanationIV}
\end{figure}

• \textbf{Experiment IV}: In the following experiment, we test a strategy that combines the optimisation of parameters with an approach based on setting an economic threshold. In the proposed approach, we discard all visits that are above a particular economic value and then, we apply a parameter optimisation on the subset as indicated in \ref{fig:ExplanationIV}. The threshold is calculated as the value that divides the dataset into two equivalent halves. Then, we discard the halve of the dataset below the threshold. In the experiment, we use two different kinds of thresholds, the first is based on the economic cost of the visits, and the second one on the expected profitability of the visits. We conducted the experiment with groups of visits as 5k, 10k, 15k... 50k.

\begin{figure}
\centering
  \includegraphics[scale=0.6]{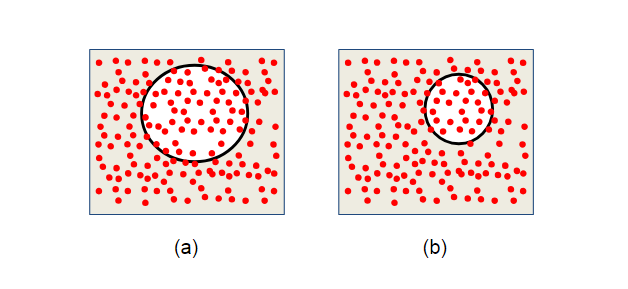}
  \caption{Increasing the searching space by looking for configurations with a smaller number of visits (b) allows to find out configurations with higher Quality Score (a).}
  \label{fig:ExplanationV}
\end{figure}

• \textbf{Experiment V}: One of the restrictions in the previous experiments was that all configurations with a number of visits lower than the required number were rejected. But, it could be the case that a campaign is offered with a number of visits slightly lower than the demanded but with average profitability that would please the advertiser.  Figure \ref{fig:ExplanationV} shows a motivating example, where the advertiser wants 15 thousand visits, and, (a) we propose a solution of 13 thousand, but, (b) with very high performance, it is likely that the advertiser may accept it as a serendipitous solution. In this experiment, we eliminate the restriction (configurations with a number of visits lower to the threshold are not discarded) which increases the search space of possible configurations to compare the performance with a scenario in which this restriction does exist. Just bear on mind that configurations with a low number of visits will still be penalised (but not rejected) as the number of visits is a factor to calculate how good a configuration is.

\section{Results and Discussion} \label{sec:results}

In this section, we discuss the results of the experiments of the proposed methodology. To measure the improvement of the proposed method, we have conducted five different experiments based on parameter optimisation combined with other ideas as discussed in the previous section. Experiments have been performed using 17 datasets, and the obtained results are the summation of the average profitability of all the datasets. The designed algorithm collects some parameters that indicate the performance of the campaign. The most important ones are shown in table \ref{tab:Solutions}.  This table shows the collected information in experiment II for the first of the 17 datasets. Other variables such as $Quality\,Score$ or the remaining size of the dataset can be calculated from these values.

\begin{table}[htbp]
\centering { \noindent\adjustbox{max width=\textwidth}{%

\begin{tabular}{|c|c|c|c|c|c|}
\hline
\textbf{Nº} & \textbf{Avg Prof} & \textbf{Nº Rows} & \textbf{Selected Cols} & \textbf{Values for each column} & \textbf{Time} \\ \hline
1 & 4667.60 & 1198 & (0, 1, 2, 4, 5, 8) & [25259032, 7477605, 28051086, 32440044, 1973606, 29196072] & 193.82 \\ \hline
2 & 4139.91 & 1199 & (0, 3, 4, 5, 7) & [25259032, 29196072, 32440044, 1973606, 23998111] & 189.89 \\ \hline
3 & 3426.97 & 1200 & (1, 2, 3, 4, 8) & [7477605, 28051086, 23549932, 32440044, 29196072] & 186.01 \\ \hline
4 & 3144.99 & 1242 & (0, 1, 2, 4, 8) & [25259032, 26597095, 477175, 32440044, 29196072] & 184.8 \\ \hline
5 & 2821.57 & 1311 & (0, 3, 4, 5, 8) & [27093701, 29196072, 32440044, 1973606, 29196072] & 189.89 \\ \hline
6 & 2946.14 & 1165 & (1, 3, 4, 5) & [7477605, 23549932, 32440044, 1973606] & 214.69 \\ \hline
7 & 2529.13 & 1340 & (0, 4, 7, 8) & [25259032, 32440044, 29196072, 29196072] & 168.6 \\ \hline
8 & 2724.94 & 1242 & (0, 1, 2, 3, 5) & [25259032, 7477605, 28051086, 29196072, 1973606] & 190.55 \\ \hline
9 & 3019.14 & 1059 & (0, 4, 7) & [25259032, 32440044, 23998111] & 182.7 \\ \hline
10 & 2112.72 & 1443 & (0, 2, 4, 5) & [28928366, 28051086, 32440044, 1973606] & 171.81 \\ \hline
11 & 2302.66 & 1251 & (0, 4, 8) & [27093701, 32440044, 29196072] & 179.44 \\ \hline
12 & 1577.02 & 1739 & (1, 3, 5, 7) & [26597095, 29196072, 1973606, 23998111] & 166.21 \\ \hline
13 & 2360.01 & 1061 & (2, 4, 5) & [29310250, 32440044, 138937] & 162.28 \\ \hline
14 & 2298.28 & 1049 & (1, 2, 3, 4, 5) & [26597095, 477175, 29196072, 32440044, 1973606] & 161.89 \\ \hline
15 & 2021.47 & 1185 & (0, 2, 3, 4, 7) & [25259032, 28051086, 23549932, 32440044, 29196072] & 153 \\ \hline
16 & 1201.57 & 1889 & (1, 2, 5) & [7477605, 28051086, 1973606] & 146.14 \\ \hline
17 & 1311.20 & 1582 & (0, 1, 4, 7) & [25259032, 26597095, 32440044, 29196072] & 150.53 \\ \hline
18 & 1852.56 & 1097 & (0, 7, 8) & [28928366, 29196072, 29196072] & 172.4 \\ \hline
19 & 1107.49 & 1653 & (1, 2, 4, 5, 7) & [7477605, 28051086, 32440044, 28928366, 29196072] & 169.35 \\ \hline
20 & 1149.95 & 1486 & (1, 2) & [26597095, 477175] & 150.95 \\ \hline
21 & 1531.85 & 1095 & (2, 4, 5, 6) & [28051086, 32440044, 32440043, 31785010] & 152.9 \\ \hline
22 & 1478.03 & 1092 & (0, 3, 5) & [25259032, 29196072, 1973606] & 139.73 \\ \hline
23 & 674.25 & 2223 & (0, 1) & [25259032, 7477605] & 124.6 \\ \hline
24 & 1229.81 & 1173 & (0, 2, 4, 7) & [27093701, 28051086, 32440044, 29196072] & 139.73 \\ \hline
25 & 1083.83 & 1243 & (3, 4) & [23549932, 32440044] & 143.41 \\ \hline
26 & 573.84 & 2070 & (0) & [28928366] & 139.38 \\ \hline
27 & 836.13 & 1237 & (1, 4, 6) & [26597095, 32440044, 31785010] & 146.03 \\ \hline
28 & 921.12 & 1067 & (0, 2, 3, 4, 5, 6) & [30763035, 6172878, 29196072, 32440044, 1973606, 31785010] & 122.21 \\ \hline
29 & 706.58 & 1373 & (1, 3, 4) & [26597095, 29196072, 32440044] & 121.61 \\ \hline
30 & 797.88 & 1196 & (4, 5, 7) & [32440044, 1973606, 23998111] & 139.58 \\ \hline
\end{tabular}
}}
\caption{Example of the fields collected by the algorithm in experiment II selecting the best 1000 impressions and eliminating them from the dataset.}
\label{tab:Solutions}
\end{table}

\subsection{Experiment I}

\begin{figure}
\centering
  \includegraphics[scale=0.6]{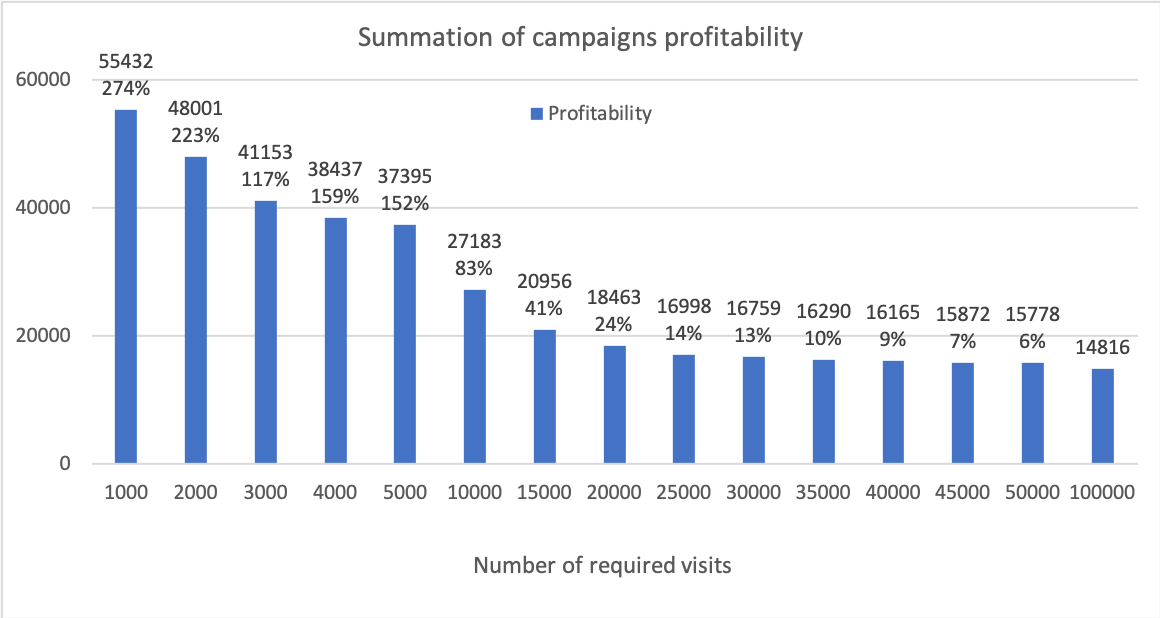}
  \caption{Experiment I confirms that as the number of required visits increases, the average profitability declines.}
  \label{fig:ExpI}
\end{figure}

The objective of this first experiment is to confirm that it is possible to enhance campaigns performance by means of parameter configuration. The results of the experiment are shown in table \ref{tab:expI}, which shows the average profitability of the best campaign with a number of visits greater than or equal to the number required by the advertiser. Figure \ref{fig:ExpI} shows the summary as a bar graph, in which it can be seen that the fewer visits are needed to make the campaign, the higher the yield becomes (the percentage indicates the increment over the average profitability). It can also be seen that the performance falls abruptly between the first five limits (between 1000 and 5000), but from that point, it tends to diminish slowly in a progressive way. We can conclude from the results that finding high-performance configurations using our methodology is feasible and that as the required number grows the profitability decreases.

\begin{table}[htbp]
\caption{In Experiment I it is calculated the configurations with the highest Average Profitability with at least the number of visits required by the advertisers.}
\begin{tabular}{|c|c|c|c|c|c|c|c|c|c|c|}
\hline
\textbf{Dataset} & \textbf{5000} & \textbf{10000} & \textbf{15000} & \textbf{20000} & \textbf{25000} & \textbf{30000} & \textbf{35000} & \textbf{40000} & \textbf{45000} & \textbf{50000} \\ \hline
1 & 4206.8 & 3244.8 & 2849.3 & 1533.5 & 1533.5 & 1478.7 & 1478.7 & 1478.7 & 1478.7 & 1478.7 \\ \hline
2 & 4152.6 & 3152.3 & 2746.1 & 2746.1 & 1545 & 1530.3 & 1530.3 & 1530.3 & 1530.3 & 1530.3 \\ \hline
3 & 2083 & 1213.9 & 1041.5 & 1041.5 & 996.1 & 996.1 & 929.4 & 929.4 & 929.4 & 897 \\ \hline
4 & 1877.5 & 986 & 856 & 839.4 & 804.9 & 804.9 & 762.8 & 762.8 & 762.8 & 701 \\ \hline
5 & 494.4 & 370.5 & 370.5 & 350.8 & 337 & 337 & 337 & 331.7 & 331.7 & 331.7 \\ \hline
6 & 480.4 & 454.9 & 345 & 344.7 & 319.7 & 319.7 & 319.7 & 319.7 & 319.7 & 319.7 \\ \hline
7 & 4939.3 & 2570.5 & 1880.8 & 1880.8 & 1880.8 & 1880.8 & 1880.8 & 1880.8 & 1774.5 & 1774.5 \\ \hline
8 & 5062.7 & 3971.8 & 1942.7 & 1942.7 & 1942.7 & 1942.7 & 1942.7 & 1942.7 & 1812.1 & 1812.1 \\ \hline
9 & 789.3 & 549 & 409.8 & 409.8 & 364.4 & 364.4 & 364.4 & 309.2 & 309.2 & 309.2 \\ \hline
10 & 2337.5 & 2063.2 & 2063.2 & 1671.3 & 1671.3 & 1548.9 & 1504.1 & 1504.1 & 1448.1 & 1448.1 \\ \hline
11 & 1004.4 & 784.3 & 605.6 & 505.2 & 505.2 & 503 & 503 & 503 & 503 & 503 \\ \hline
12 & 2038.2 & 1876.4 & 1240.3 & 1183.8 & 1183.8 & 1183.8 & 868.6 & 868.6 & 868.6 & 868.6 \\ \hline
13 & 1161 & 511.6 & 511.6 & 511.6 & 511.4 & 511.4 & 511.4 & 483.5 & 483.5 & 483.5 \\ \hline
14 & 1274.9 & 1184.5 & 1058.6 & 774.3 & 774.3 & 746.7 & 746.7 & 746.7 & 746.7 & 746.7 \\ \hline
15 & 1820.3 & 1785.5 & 1062.4 & 1039.1 & 939.3 & 921.9 & 921.9 & 921.9 & 921.9 & 921.9 \\ \hline
16 & 1470.3 & 1197 & 1188.3 & 903.6 & 903.6 & 903.6 & 903.6 & 867.5 & 867.5 & 867.5 \\ \hline
17 & 2202.2 & 1267.1 & 784.4 & 784.4 & 784.4 & 784.4 & 784.4 & 784.4 & 784.4 & 784.4 \\ \hline
\end{tabular}
\label{tab:expI}
\end{table}

\subsection{Experiment II}

\begin{figure}
\centering
  \includegraphics[scale=0.5]{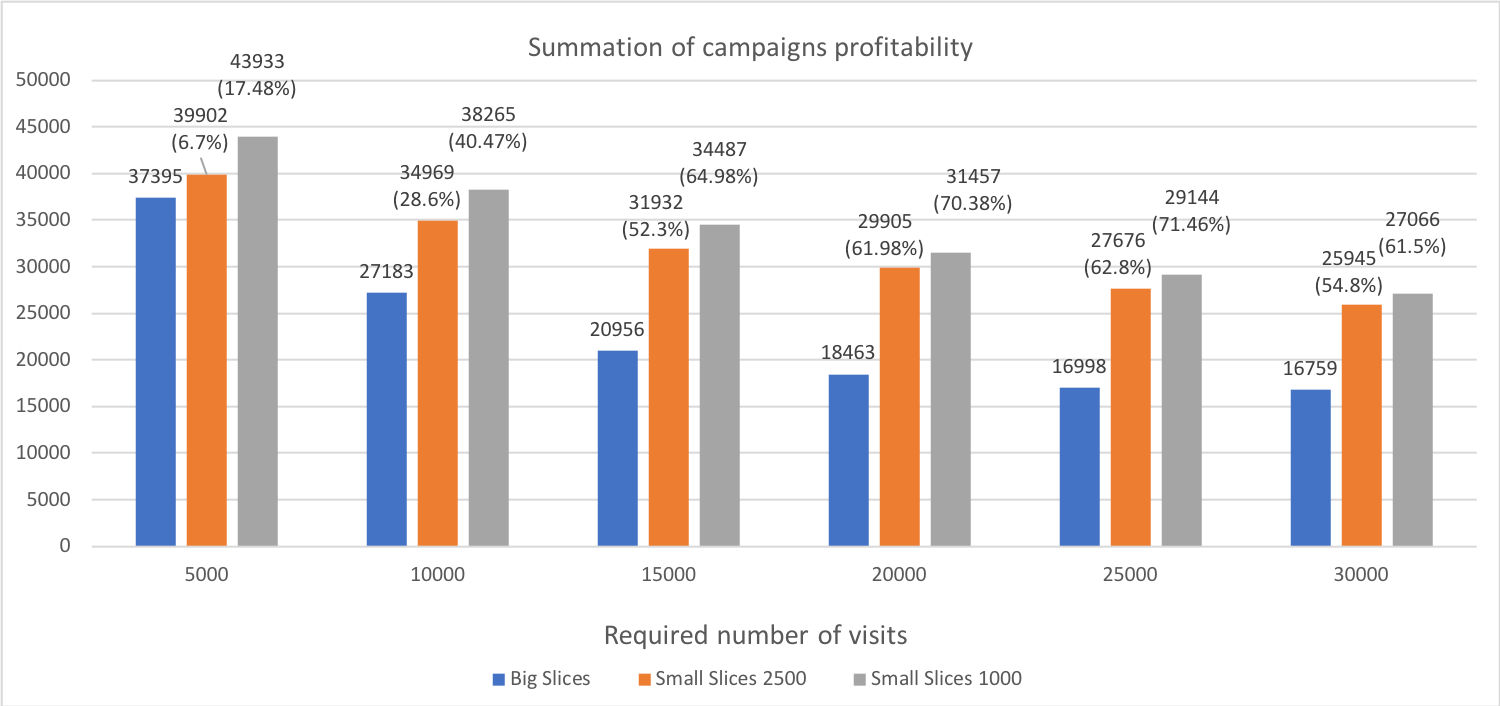}
  \caption{The technique used in Experiment II consists of selecting multiple configurations with small visits. It performs better since it is more difficult to find large groups with high performance.}
  \label{fig:ExpII}
\end{figure}

This experiment is conducted to find out if the technique of selecting a group of configurations with a small number of visits has a higher performance than selecting a single configuration with a large number of visits, where the combination of the visits of the group has a similar number of visits to the single configuration. Figure \ref{fig:ExpII} clearly shows that when the slides are of 1,000 visits, the increase in profitability is very sensitive. In addition, it can be appreciated that as the number of required visits increases, the performance when applying this technique has much better results. However, when the slides are of 2,500 visits, the improvement is very small and as the number of required visits increases, the improvement decreases. The drawback of this strategy is that it requires a higher processing cost. For example, when searching for sets of 10,000 items, the proposed strategy will search for four sets of 2,500, which requires to execute the algorithm four times instead of one. However, the improvement of the profitability confirms that it is an efficient approach.

\subsection{Experiment III}

\begin{figure}
\centering
  \includegraphics[width=0.7\linewidth]{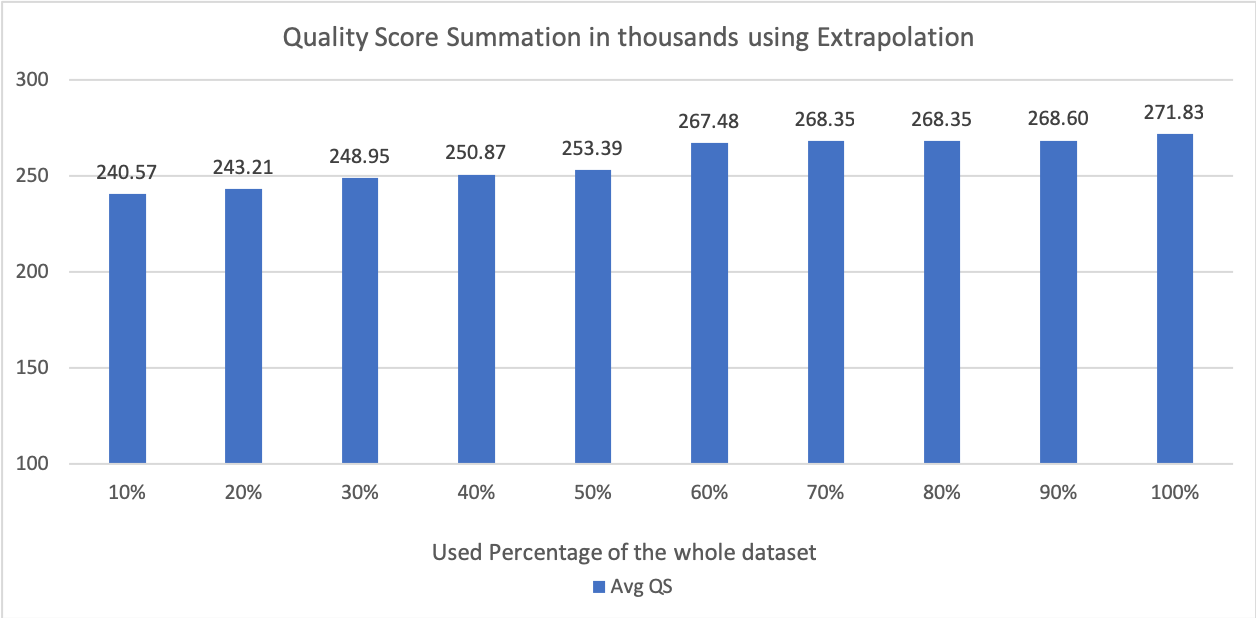}
  \caption{Extrapolating configurations from a small sample, as shown in Experiment III, is a very effective technique to find good configurations without having to do an excessive expense of previous.}
  \label{fig:ExpIII}
\end{figure}

As shown in figure \ref{fig:ExpIII}, extrapolating is an interesting way to save campaign expenses (as discussed in the previous section). The experiments show that by extrapolating from the first 10\% of the campaigns visits, it is possible to find out very profitable configurations. In particular, these configurations have an average fitness value only 8\% lower than the optimal solution. This finding is interesting in order to save economic resources, as it can be assumed that a configuration that performs well today will continue to do so on the following days.

\subsection{Experiment IV}

\begin{figure}
\centering
  \includegraphics[scale=0.6]{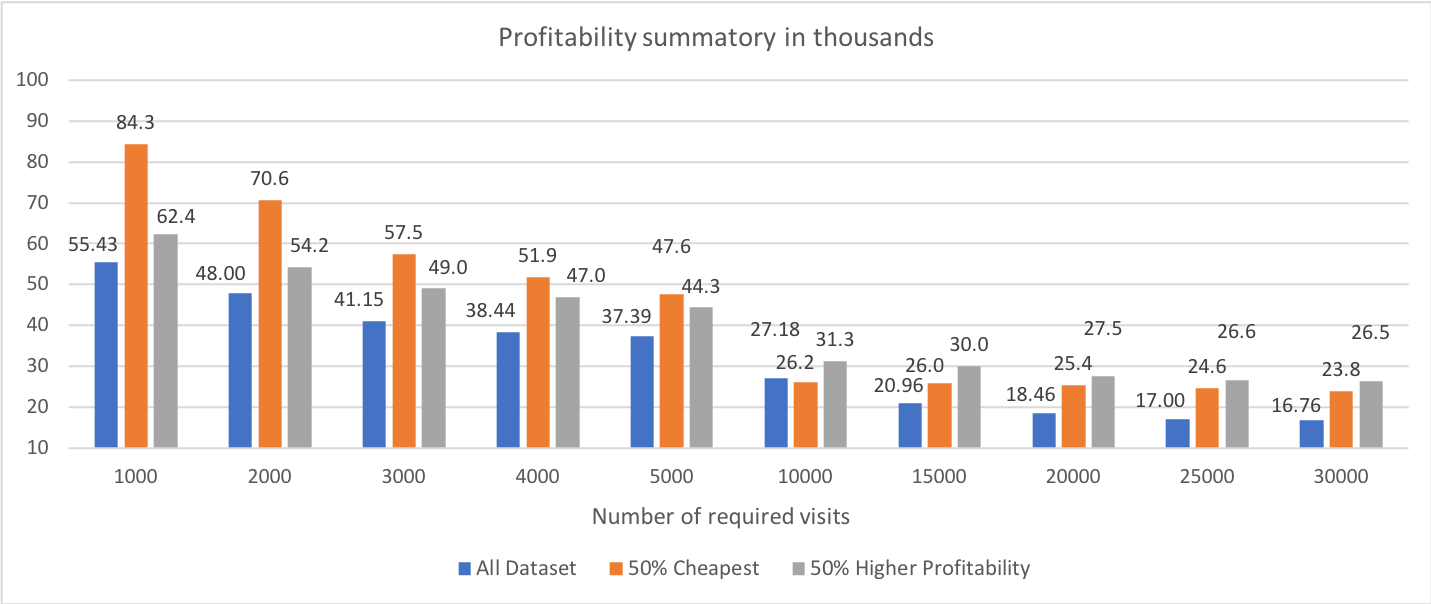}
  \caption{Experiment IV confirms that discarding visits below a certain price or profitability significantly improves the performance of the selected parameters. }
  \label{fig:ExpIVa}
\end{figure}

\begin{figure}
\centering
  \includegraphics[scale=0.6]{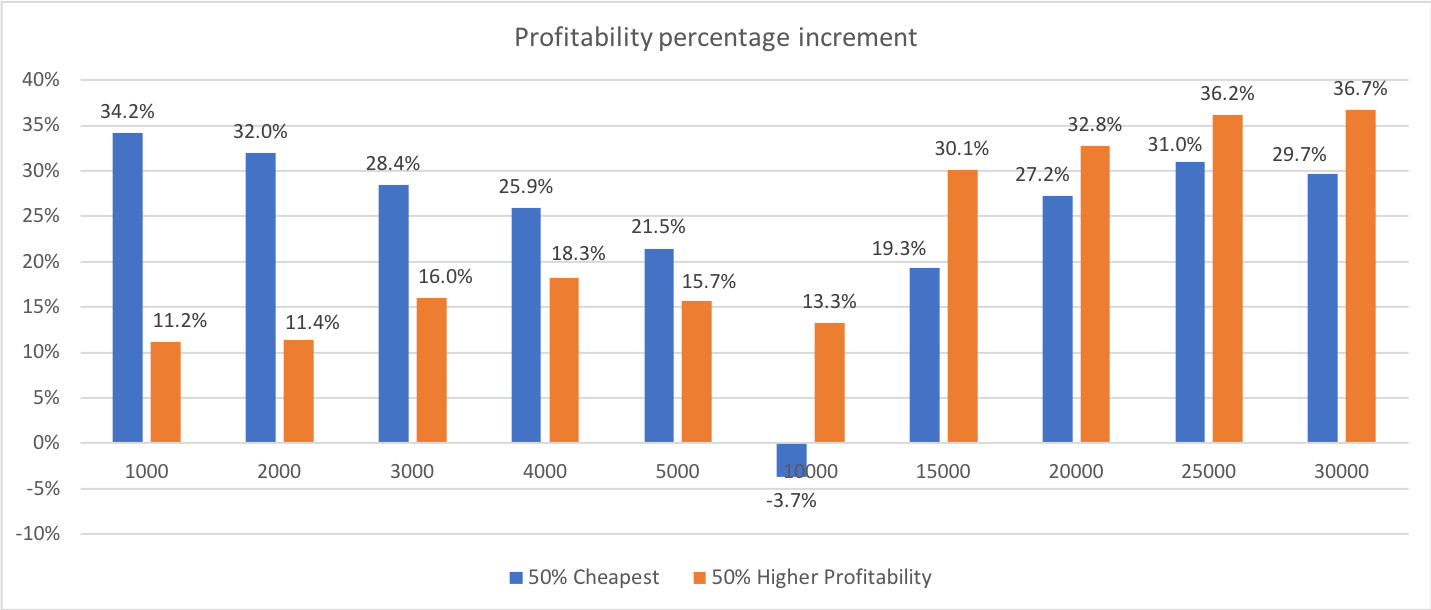}
  \caption{The performance of the approach based on the price is better until reaching 10k visits, from that point on it is better to use the threshold based on profitability.}
  \label{fig:ExpIVb}
\end{figure}

The results of experiment IV confirm that combining the selection of parameters with the application of a threshold (either by profitability or by price) increases the efficiency of the campaigns. Figure \ref{fig:ExpIVa} shows that until reaching the point of the 10k visits, using the price as a threshold is better than using the one based on profitability. Figure \ref{fig:ExpIVb} shows the increments of improvement with each threshold. However, when the number of required visits is higher than 10k, it is better to use the threshold based on profitability. We also see that as the number of required visits increases, the performance of the three approaches tends to equalise.

\subsection{Experiment V}

\begin{figure}
\centering
  \includegraphics[scale=0.5]{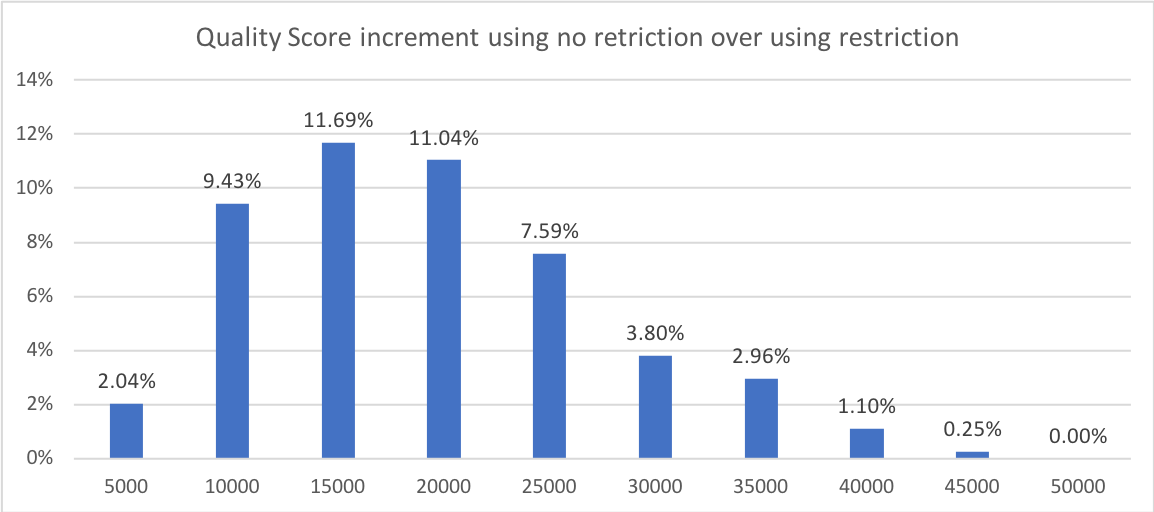}
  \caption{Experiment V shows that configurations in the search space below the threshold can be very cost efficient, low cost and with a number of visits close to the required number.}
  \label{fig:ExpV}
\end{figure}

Increasing the search space by allowing the algorithm to explore configurations with a number of visits below-the-limit allows to improve the campaign performance and decreases the economic costs of the campaign.  However, it requires a higher computational cost since the search space for the solutions is larger. As it can be seen in figure \ref{fig:ExpV}, this technique brings a substantial improvement within the range of 10k to 25k, but from 40k the improvement becomes marginal.

\subsection{Performance comparison with the methods}

Next, we compare the performance of our methodology with some of the state-of-the-art methods. Although we have not found other researches focused on increasing the performance of campaigns by means of parameter optimisation, we would like to highlight some recent publications aimed at improving the performance of RTB campaigns.

Zhang et al. \cite{zhang2014real} affirm that there are not many publications related to RTB because until 2013 (the year in which the Chinese RTB company iPinyou decided to make public some of its campaigns) there were no databases related to RTB campaigns. This same author performed a comparative study on the performance of bidding strategies applied to RTB. The results are shown according to a key performance indicator (KPI) defined as the summation of clicks and visits. Results indicate that the algorithm called LIN (Linear-form bidding of predicted CTR) improves the performance of the bidding model below max (MCPC) by 204.13\% when using $1/32$ of the budget, 24\% when using $1/8$ of the budget, and 8.5\% when using $1/2$ of the budget.

In a similar way, Zhang et al. \cite{zhang2014real} developed an algorithm to increase the number of clicks in the campaigns. In this research, it compares the performance of a new method called ORTB (Optimal real-time bidding) with LIN. The average increment of the summation of clicks for the nine campaigns using different budgets is 84.3\% when the budget is 1/64, 28.61\% when the budget is 1/32, 16.14\% for 1/16, 9.19\% for 1/8, 4.43\% for 1/4, and 1.94\% for 1/2 \cite{zhang2014optimal}. 

Another interesting publication is that of Lee et al. \cite{lee2013real} in which an adaptive algorithm to select quality impressions is presented. The algorithm takes into account the performance of previously displayed impressions while distributes the budget evenly overtime to reach the widest possible audience. In CPM flat campaigns, the average CTR increment for seven campaigns was 123.7\%. The performance of this methodology was also evaluated using ten dynamic CPM campaigns and the increment in performance with respect to conversions (CPA) and the number of clicks (CPC) was 30.9\% and 19.0\% respectively. 

We also find relevant the publication of Do et al. \cite{du2017improving} in which they improved the performance of the RTB through a Constrained Markov Decision Process (CMDP) based on a reinforcement learning framework. A distributed representation model is used to estimate the CTR value where the estimated CTR is the state, and the price of the action and the clicks are the reward. We see that the CMDP performance in terms of the number of clicks for the sum of ten campaigns is 12.6\% better and the expected cost-per-click (eCPC) is 9.13\% lower than in \textit{Sparse Binary} which was considered as a baseline.

Shioji et al. \cite{shioji2017neural} used neural embedding strategies like word2vec to improve the estimation of the users' response to the displayed adverts (a.k.a. CTR). The Word2vec technique is used to learn distributed representations from the internet browser history of the users. This approach is able to improve the accuracy of the CTR estimations results as follows: 4.90\% for 0.3k, 3.39\% for 1k, 2.87\% for 10k, and 1.38\% for 100k, where the first number of the tuple indicates the side of the training data and the second the AUC improvement.

\section{Conclusions and Future Work}

RTB platforms are becoming a very beneficial advertising model for publishers and advertisers. It is no wonder that the estimated volume of impressions managed by RTB networks will continue to increase in the coming years. It does not seem unreasonable to say that in the near future RTB can replace advertising networks platforms or at least take a significant market share.

In this paper, we propose a novel methodology based on parameter configuration to find profitable campaigns for advertisers in an automatic fashion. In this sense, we think that the proposed approach is interesting because, to our knowledge, it covers a gap in RTB campaign optimisation research.

The developed experiments prove that the presented methodology improves the results of RTB campaigns in a substantial way. Not only that, the combination of parameter optimisation with other approaches such as small campaign selection, setting a threshold, configuration extrapolation, or increasing the solution search space, improves, even more, the obtained results. However, these results may vary depending on several variables of a campaign such as the target, the moment when it is launched, or the behaviour of the rest of the competitors.

Implementing the methodology could enable RTB platforms as well as advertising networks that manage third-party campaigns to suggest advertisers better configurations for their campaigns. Profitable campaigns will eventually boost the performance of advertisers companies, making RTB a more attractive platform, which in turn will make RTB advertisers more willing to launch more campaigns.

It could seem that as the selected number of configurations increases, the gain of the average profit by campaign becomes trivial. But, here, we argue that, first, in RTB many ad networks coexist with their advertisers, therefore, when a platform decides not to bid on a particular impression, it does not imply that it is lost, but instead, it will be disputed among the rest of the advertisers. Additionally, it may be the case that some impressions may have low conversion probability for an advertiser but a high probability for another advertiser as it depends on the nature of the advert. Secondly, there are two types of campaigns: ones based on branding and others based on performance. Impressions not valuable from a performance-based perspective could be valuable for branding-based campaigns; where the goal is to increase the brand value instead of looking for profits in the short term.

As future work, it could be a good idea to combine several techniques. For example, setting an economic threshold with small campaign selection and increasing the search space. The new methodology could be tested in different scenarios with other payment methods such as pay-per-click or pay-per-acquisition. Using algorithms to find a suboptimal solution but in less time could also be a good starting point. To this end, multi-objective evolutionary algorithms such as NSGA-II (Non-dominated Sorted Genetic Algorithm) could be a good solution.

\bibliographystyle{apalike} % apalike unsrt
\bibliography{biblio}

\end{document}